# MACHINE LEARNING TRAINING OPTIMIZATION USING THE BARYCENTRIC CORRECTION PROCEDURE

Sofía Ramos-Pulido, Neil Hernández-Gress and Héctor G. Ceballos-Cancino

Tecnologico de Monterrey, Av. Eugenio Garza Sada 2501 Sur, Tecnológico, 64849 Monterrey, N.L

## ABSTRACT

*Machine learning (ML) algorithms are predictively competitive algorithms with many human-impact applications. However, the issue of long execution time remains unsolved in the literature for high-dimensional spaces. This study proposes combining ML algorithms with an efficient methodology known as the barycentric correction procedure (BCP) to address this issue. This study uses synthetic data and an educational dataset from a private university to show the benefits of the proposed method. It was found that this combination provides significant benefits related to time in synthetic and real data without losing accuracy when the number of instances and dimensions increases. Additionally, for high-dimensional spaces, it was proved that BCP and linear support vector classification (LinearSVC), after an estimated feature map for the gaussian radial basis function (RBF) kernel, were unfeasible in terms of computational time and accuracy.*

## 1. INTRODUCTION

Artificial neural networks and machine learning models have assumed paramount significance in numerous applications that profoundly affect human activities. These methodologies have gained prominence due to their exceptional predictive accuracy [1-6]. Neuronal networks, now named deep learning, re-emerged after 2010 due to massive improvements in computer resources, some innovations, and successful applications [7]. Support vector machines and gradient boosting stand as pillars in the field of machine learning [8-11].

Deep learning has emerged as a cornerstone for tackling intricate tasks such as object recognition, speech recognition, and the development of autonomous vehicles [12]. Support vector machines have demonstrated their efficacy in domains like computer security, image categorization, and the extraction and recognition of soft biometrics [13]. Meanwhile, gradient boosting has found resounding success in a wide array of sectors, including finance [14] education [15], and the cryptocurrency realm [16], among countless others.

In addition to their remarkable accuracy and proven success in various applications, Support Vector Machines (SVM) are renowned for their ability to construct intricate decision boundaries, even when dealing with datasets featuring only a limited number of features [17]. One compelling theoretical aspect that distinguishes SVM from other algorithms is its convex objective function, guaranteeing the discovery of the optimal solution consistently [18]. Additionally, SVM possesses the unique characteristic of architecture self-determination, eliminating the need for prior definition.

Turning to the Gradient Boosting algorithm, as discussed in the work of [19], tree-based ensemble methods, like gradient boosting, offer interpretable outcomes with minimal data preprocessing requirements. The boosting family of algorithms has consistently ranked among the most accurate classifiers across a wide range of datasets [20]. Despite its sensitivity to noise and outliers, particularly in smaller datasets, it consistently exhibits lower testing error rates [20].

There is no denying the paramount importance, efficiency, and success stories associated with algorithms like neural networks, support vector machines, and gradient boosting, even when dealing with small databases. However, it is crucial to acknowledge that despite their many merits, all three methods face their share of challenges.

Deep learning, which employs backpropagation (BP), has encountered criticism due to certain theoretical limitations. BP does not offer a guarantee of reaching the absolute minimum [21], and the necessity to predefine the model's architecture poses a significant hurdle. As noted by [17], neural networks are highly sensitive to parameter choices, and in the case of large networks, they incur substantial memory usage and slow training and testing times.

Support vector machines, on the other hand, have faced a potential impediment to widespread adoption in the form of execution time and memory requirements. Handling SVM on datasets exceeding 100,000 entries or more can prove challenging, particularly concerning runtime and memory consumption [17].

Meanwhile, gradient boosting is not immune to computational challenges. High memory consumption, slower evaluation speeds, and time-intensive processes become evident when dealing with considerably large ensembles [22].

The primary objective of this study is to address the challenge of extensive execution times when dealing with large datasets in the context of methodologies such as support vector machines, neural networks, and gradient boosting. While substantial progress has been made in addressing this issue in the past, it remains a pressing concern, particularly given the ever-expanding volume of data generated by industry and scientific endeavors.

To tackle this challenge, we propose leveraging a previously introduced approach that involves the initialization of these algorithms with a barycentric correction procedure (BCP). This procedure aims to identify a subset of instances refined within the training set to enhance the execution of the above-mentioned algorithms. BCP is grounded in geometric principles initially proposed by [23], offering faster convergence than the Perceptron in linear cases [23].

We hypothesize that by working with a training set comprising fewer but more strategically selected instances, we can significantly reduce both the time and memory requirements while preserving high levels of accuracy.

## 2. METHOD

This section shall present the conceptual framework of the proposal, which initiates with the application of the BCP algorithm. It will discuss the hypothesis about reducing time and expound on the basics of BCP; subsequently, the subsection dedicated to experiments will provide the process for synthetic data generation. We will elucidate the experiments encompassing both synthetic and real datasets. The real dataset employed in our study comprises educational information with features that impact graduates and educational institutions.

## 2.1. Proposal

The proposal is exemplified in Figure 1. The algorithm consists of the following three steps:

- Run BCP on the complete data set and generate an approximate hyperplane to the solution.
- Extract a subset of cases close to the hyperplane generated by BCP.
- Run SVM, gradient boosting (GB), or neuronal networks on the reduced set.

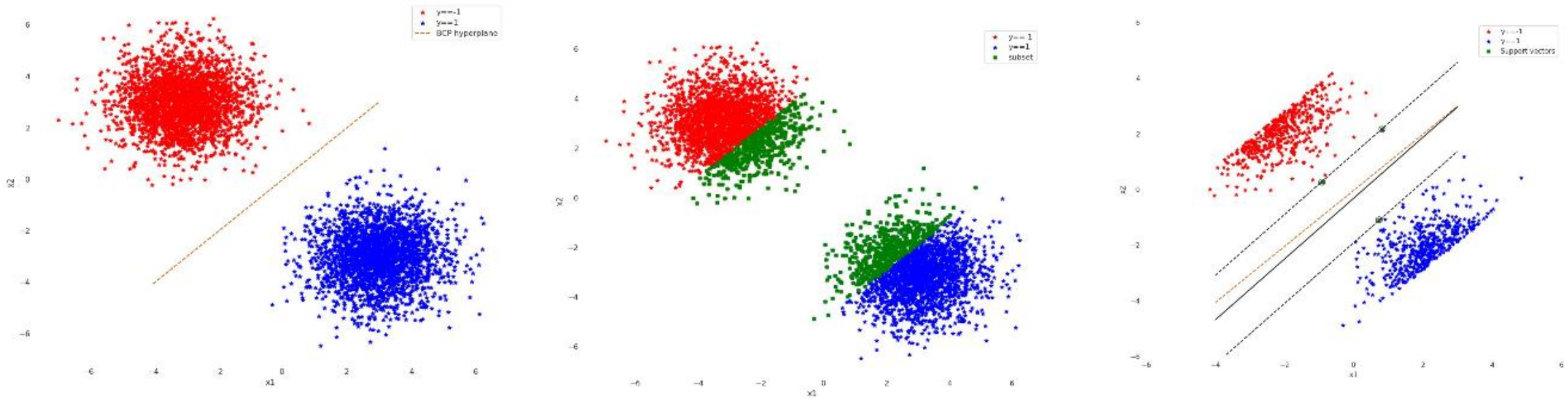

Figure 1. Stages of the Proposal that combines BCP and SVM. Left: Run BCP; Center: Extract the subset; Right: Run SVM in the reduced set.

Since the algorithms will be executed on a smaller data set, lower memory requirements are hypothesized. Furthermore, we hypothesize that in higher dimensional cases, BCP will guide the algorithms with a reduction in execution time.

## 2.2. Barycentric Correction Procedure

The barycentric Correction Procedure used in step 1 of the proposal was stated in [23]. BCP depends on calculating separate weights and a threshold. The training is done by an iterative correction of the weights of the barycenters allowing to minimize the number of misclassified values. The algorithm defines a hyperplane $w^t x + \theta$dividing the input space into two classes. First, let us define $I_1 = 1, .., N_1$,and $I_0 = 1, .., N_0$, where $N_0$ is the number of negative cases, and $N_1$ is the number of positive cases. The following weighted averageare the barycenters of the classes defined by [23],

$$b_1 = \frac{\sum_{i \in I_1} \alpha_i x_i}{\sum_{i \in I_1} \alpha_i}, \qquad b_1 = \frac{\sum_{i \in I_0} \mu_i x_i}{\sum_{i \in I_0} \mu_i} \tag{1}$$

where $\alpha_i$ and $\mu_i$ are weighting coefficient vectors of size $N_1$ and $N_0$, respectively.
The weight vector w is a vector difference $w = b_1 - b_0$. The bias term, $\theta$, is computed with,

$$\theta = \frac{\max \gamma_1 + min\, \gamma_0}{2} \tag{2}$$

where $\gamma(x) = -wx$, $\gamma_1 = \{\gamma(x_i) | x_i \in \text{positive class}\}$ and $\gamma_0 = \{\gamma(x_i) | x_i \in \text{negative class}\}$. The barycentric correction is calculated using the modification in the weighting coefficients.

$$\alpha_{new} = \alpha_{old} + \beta, \quad \mu_{new} = \mu_{old} + \lambda \tag{3}$$

where $\beta = min\{1, max[30, N_1/N_0]\}$ and $\lambda = min\{1, max[30, N_0/N_1]\}$, [24]. In some cases, BCP was shown to be 70,000 faster than a perceptron [24].

## 2.3. Experiments

The proposal is validated in synthetic samples with different features and instances and in a dataset of educational information property of a university. The experiments were conducted with the default resources of Colab, the free and serverless Jupyter Notebook that executes Python code [25].

We proved the proposal with linear separable and non-linear separable synthetic data. Linearly and non-linear separable synthetic samples were generated with different numbers of instances (n) and various dimensions or features (p).

*Linear separable synthetic data procedure*

- First, generate n samples of p features, $X = [x_1, \dots, x_n]$, with random samples from a standard normal distribution.
- Compute the weighting vector of dimension $(1, p)$, $\beta$, with random samples from a standard normal distribution.
- Calculate$z$, the matrix product of $\beta$ and $X, z = \beta X$.
- Last, when $z < 0$ corresponds to class -1 and class 1 was defined when $z > 0$.

The left side of Figure 2 shows an example in two dimensions and 5,000 instances. The red points represent one population, and the blue points represent another.

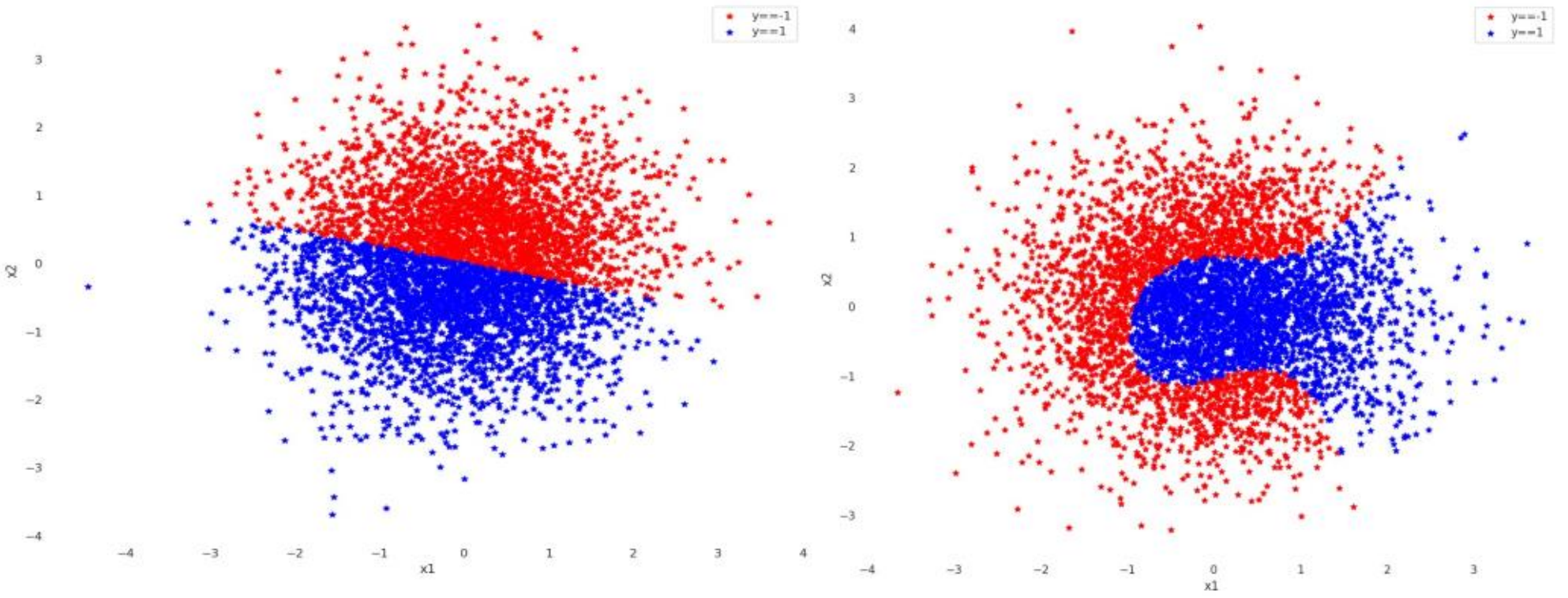

**Figure 2.** Synthetic data

*Nonlinearly separable case data*

1. First, generate n samples of p features, $X = [x_1, \dots, x_n]$, with random samples from a standard normal distribution.
2. From the previous inputs, generate inputs in a higher dimensional space $x^*$.
3. Calculate the weighting vector of dimension $(1, p)$, $\beta$, with random samples from a standard normal distribution.
4. Finally, generate classes for each sample as follows: $y_i = -1$ if $x^{*T}\beta < 0$ and $y_i = 1$ if $x^{*T}\beta > 0$.

We used a polynomial of degree p to generate inputs in a high dimensional space. The data set is linearly separable in the transformed inputs but not necessarily in the original $x's$. For example, if p = 2 and we transform the $x$ as $x^* = (x_1, x_2, x_3, x_4)$, then generating labels as described above would generate linearly separable data if $\beta_2 = 0$ since $x_i^T \beta = x_{i1}\beta + x_{i2}\beta$.

The right side of Figure 2 shows the linear non-separable synthetic data generated in two dimensions and 5,000 instances following the last algorithm.

### 2.4. Real Data

Finally, the proposal was validated with real educational data property of on university. The University, on the 80th anniversary, surveyed the alumni to know their social and economic impact. The QS Intelligence Unit Team and researchers from the University approached the descriptive analysis from this survey, and the University owns a report [26]. For this study, the University provided the dataset without personal identification. The questionnaire contains more than 50 questions, almost all of multiple options. For this specific study, we do not want to know the associated input features with the output target; we want only to predict the following output features:

**Satisfaction** If an alumni would study again at the University.
**Salary** If the alumni have a salary bigger or not than the median.
**CEO** If an alumni have a management CEO position.

The input features include the following and more: age, gender, school, campus, level of education, current address, region of birth, parent's education, parent's occupation, working hours, years of working abroad, life satisfaction, income satisfaction, evaluations in social intelligence, self-knowledge management, communication. The total number of features after dummies were created for nominal features is 104, and the number of instances is 25,359.

## 3. RESULTS

### 3.1. Kernel approximation

The synthetic data used for the experiments in this section was created with the algorithms presented in the method section with d = 3 and 10 dimensions. As a reference, the left side of Figure 2 shows the case of d = 3 in two dimensions.

Previously, studies have examined an estimated feature map for the Gaussian radial basis function (RBF) kernel, revealing that the approximate feature map results in significant speed improvements with minimal impact on classification accuracy [27]. The approximate feature map of RBF also proves effective with BCP in low-dimensional space.

However, BCP or linear support vector classification (LinearSVC) after the RBF kernelfeature mapapproximation (RBFsampler) is far for outperform the exact SVM evaluations in accuracy. Several experiments were implemented with synthetic nonlinear separable data in ten dimensions, with the maximum number of components (possible with the computational resources) in the approximation (around 4,000) and with different iterations. Table 1 shows the average performance observed in various experiments versus the machine learning algorithms.

**Table 1.** Accuracy of the kernel approximation

| | 10 features | | | | |
|---|---|---|---|---|---|
| n | RbfSampler + BCP | RbfSampler+ LinearSVC | SVM | NNK | GB |
| 5e + 04 | 67 | 67 | 97 | 93 | 88 |
| 1e + 05 | 66 | 66 | 98 | 94 | 89 |

*RbfSampler + BCP: BCP after the RBF kernel feature map approximation, RbfSampler + LinearSVC: linear support vector classification after the RBF kernel feature map approximation, SVM: support vector machine, NNK: neuronal networks, GB: gradient boosting.*

As indicated by the findings presented in Table 1, surpassing the performance of precise SVM evaluations through kernel approximations proves unfeasible in terms of computational time and accuracy. Consequently, an alternative course of action must be pursued for more than ten features. The subsequent two sections elucidate the outcomes of the proposed approach when applied to synthetic and real data.

## 3.2. Synthetic Data

### Linear separable case

Table 2 shows the outcomes of the proposal with linear separable synthetic data versus the result of SVM, neuronal networks, and gradient boosting in several instances and 10 and 20 dimensions. The left side of Figure 1 shows the first two steps of the algorithm in a two-dimensional linearly separable case. The separating equation with the BCP is indicated in orange, and the green dots indicate the reduced set to which SVM is applied. The third step of the algorithm is depicted on the right side of Figure 1, the highlighted points on the black dotted lines indicate the support vectors, and the solid line represents the hyperplane generated by SVM on the reduced set.

**Table 2.** Linear separable experiments

| | | 10 features | | | | | | |
|---|---|---|---|---|---|---|---|---|
| n | | BCP | BCP + SVM | SVM Lineal | BCP + NNK | NNK | BCP + GB | GB |
| 5e + 04 | Time | 3 s | 4 s | 4 s | 25 s | 17 s | 1 s | 1 s |
| | Accuracy | 100 | 100 | 100 | 100 | 100 | 97 | 97 |
| 1e + 05 | Time | 10 s | 7 s | 14 s | 21 s | 56 s | 2 s | 2 s |
| | Accuracy | 100 | 100 | 100 | 100 | 100 | 98 | 98 |
| 5e + 05 | Time | 28 s | 123 s | 264 s | 94 s | 183 s | 5 s | 7 s |
| | Accuracy | 100 | 100 | 100 | 100 | 100 | 98 | 98 |
| | | 20 features | | | | | | |
| 5e + 04 | Time | 26 s | 7 s | 6 s | 27 s | 22 s | 2 s | 2 s |
| | Accuracy | 100 | 100 | 100 | 100 | 100 | 94 | 94 |
| 1e + 05 | Time | 60 s | 11 s | 26 s | 33 s | 60 s | 4 s | 3 s |
| | Accuracy | 100 | 100 | 100 | 100 | 100 | 95 | 95 |
| 5e + 05 | Time | 376 s | 216 s | 492 s | 81 s | 167 s | 8 s | 9 s |
| | Accuracy | 100 | 100 | 100 | 100 | 100 | 96 | 96 |

*BCP + SVM: Proposal that combines BCP and SVM, BCP + NNK: Proposal that combines BCP and NNK and BCP + GB:Proposal that combines BCP and GB.*

The advantages in the linearly separable case are in terms of memory and time. When the number of instances is 500,000, and the number of features is ten or more, our proposal is far faster than SVM, neuronal networks, and gradient boosting alone. In fact, the proposal is faster than SVM and neuronal networks since there are 100,000 instances and more than ten features; see Table 2. Additionally, in memory, for example, when the number of dimensions is 20, and the number of instances is 15 million, it is impossible to run SVM using the default memory resources of Google Colab; however, with the proposal, it is possible to run the algorithm in this case.

### Nonlinear separable experiments

Table 3 shows the outcomes of the proposal with nonlinear separable synthetic data versus the result of SVM, neuronal networks, and gradient boosting in several instances. The synthetic data used for the experiments in this section was created with the algorithms presented in the method section with d = 3 and 10 or 20 dimensions.

**Table 3.** Linear separable experiments

| | | | | | | | |
|---|---|---|---|---|---|---|---|
| | **10 features** | | | | | | |
| **n** | | **BCP + SVM** | **SVM Lineal** | **BCP + NNK** | **NNK** | **BCP + GB** | **GB** |
| 5e + 04 | Time | 29 s | 51 s | 48 s | 64s | 3 s | 1 s |
| | Accuracy | 96 | 97 | 92 | 93 | 88 | 88 |
| 1e + 05 | Time | 118 s | 180 s | 86s | 125 | 4 s | 2 s |
| | Accuracy | 97 | 98 | 93 | 94 | 89 | 89 |
| 5e + 05 | Time | 2799 s | 4170 s | 421 s | 561 | 9 s | 7 s |
| | Accuracy | 99 | 99 | 94 | 95 | 89 | 89 |
| | **20 features** | | | | | | |
| 5e + 04 | Time | 77 s | 132 s | 52 s | 63 s | 4 s | 2 s |
| | Accuracy | 89 | 90 | 81 | 82 | 75 | 75 |
| 1e + 05 | Time | 320 s | 522 | 111 s | 134 s | 5 s | 3 s |
| | Accuracy | 93 | 94 | 84 | 85 | 75 | 75 |
| 5e + 05 | Time | 5380 s | 8986 s | 373 s | 621 s | 16 s | 10 s |
| | Accuracy | 98 | 98 | 86 | 87 | 75 | 75 |

*BCP + SVM: Proposal that combines BCP and SVM, BCP + NNK: Proposal that combines BCP and NNK and BCP + GB:Proposal that combines BCP and GB.*

Time reduction is the advantage of our proposal with the nonlinearly separable synthetic data. When the number of instances is 50,000 or more, our proposal is far faster than SVM alone and faster than neuronal networks. However, our proposal needs to improve the time of xgboost because this algorithm is very efficient even with millions of data.

It is important to mention that the improvement of time using our proposal is not affecting the accuracy. The accuracy in the experiments with linear separable and nonlinear separable synthetic data was exceptional in both the proposal and the machine learning algorithm.

### 3.3. Real Data

Table 4 shows the outcomes of the proposal and the machine learning algorithms in three target variables from real educational data. Because the three target variables are unbalanced, we present accuracy and AUC.

**Table 4.** Real data

| | **Target: Satisfaction** | | | | | |
|---|---|---|---|---|---|---|
| **n** | **BCP + SVM** | **SVM** | **BCP + NNK** | **NNK** | **BCP + GB** | **GB** |
| Accuracy | 87 | 86 | 79 | 86 | 87 | 87 |
| AUC | 70 | 67 | 72 | 74 | 76 | 75 |
| Time | 23 s | 71 s | 45 s | 81 s | 10 s | 9 s |
| | **Target: CEO** | | | | | |
| Accuracy | 85 | 85 | 88 | 88 | 90 | 91 |
| AUC | 54 | 54 | 73 | 64 | 77 | 80 |
| Time | 29 s | 62 s | 58 s | 123 s | 19 s | 8 s |
| | **Target: Salary** | | | | | |
| Accuracy | 72 | 73 | 75 | 77 | 80 | 81 |
| AUC | 69 | 69 | 74 | 74 | 80 | 80 |
| Time | 90 s | 146 | 93 s | 114 s | 22 s | 10 s |

*BCP + SVM: Proposal that combines BCP and SVM, BCP + NNK: Proposal that combines BCP and NNK and BCP + GB:Proposal that combines BCP and GB.*

Time reduction without affecting the metrics was observed between the proposal and SVM for the three target variables. There was no advantage found over the gradient boosting xgboost algorithm. There is an advantage in the time of our proposal versus neuronal networks, but it affected the accuracy and AUC in the target variable of satisfaction.

Additionally, because SVM's metrics are below the other algorithms, tunning is necessary, implying more execution time. In some experiments, we found that we needed approximately six hours to implement Bayesian optimization with SVM. However, if our proposal is implemented, the time needed is reduced to one hour or less, achieving at least the values in the metrics observed in gradient boosting and neuronal networks.

## 4. CONCLUSIONS

The combination of the BCP and SVM can solve SVM memory and time problems in high-dimensional linear and nonlinear separable classification problems. The combination of BCP and SVM is much faster than SVM alone. These proposal's SVM improvements were tested with synthetic and real data without affecting the accuracy or AUC. Furthermore, combining BCP and neuronal networks also helps neuronal networks with time problems in high-dimensional classification problems.That improvement of the proposal with neuronal networks is validated with synthetic data.

Most of the experiments were conducted with default parameters in the algorithms; however, in some experiments, we notice that when we change to kernel poly, the time increases drastically using SVM alone, and the proposal helps much more in this case. Also, when tuning is needed, which is the case of SVM in experiments with real data, finding the best parameter to achieve at least the same accuracy of neuronal networks implies much time with SVM. However, tunning with the combination of BCP and SVM consumes much less time than SVM alone and achieves the same accuracy as SVM with tunning.

To conclude, the proposal minimizes the long execution time of neuronal networks and SVM, emphasizing that it is SVM that can benefit the most from the proposal. In the general case, we can recommend our proposal in high-dimensional training sets with more than 50,000 instances and more than 10 features.

## AUTHORS

**Sofía Ramos-Pulido**is currently pursuing a Ph.D. in Computer Science at the Tecnologico de Monterrey. She boasts over five years of experience as a professor, specializing in Probability, Statistics, Mathematics, and Stochastic Processes. Her research focuses on Data Science, Statistics, Machine Learning, and Deep Learning, driven by a keen desire to apply her findings for a positive social impact.

**Neil Hernández-Gress** BS´93, MSc ´95 and PhD ´98, is Associated Vice Rector for Research at Tec de Monterrey. His research interests are: Data Science, Machine Learning, Neural Networks and several aspects of Artificial Intelligence. For the last 20 years, he has developed theoretical methods and applied methodologies for a number of applications in engineering, finance and health. He is Professor of Artificial Intelligence, Neural Networks, Data Science and Innovation methodologies for more than 20 years. He is member of the National Research Council since 2000. He is the National Contact Point (NCP-ICT) in Mexico named by CONACYT and the European Commission. He is the author of more than 50 Research Papers and the recipient of funded research projects in more than 20MUSD. He is also an advocate for innovation as a resource for wealth creation.

**Héctor G. Ceballos-Cancino** is director of the Living Lab & Data Hub of the Institute for the Future of Education at Tecnológico de Monterrey. Previously, he was head of the Scientometrics office at the Tec's Research Vice-Rectory. He has a master and a PhD in Intelligent Systems by Tec de Monterrey. Hector Ceballos is also ascribed to the research group on Advanced Artificial Intelligence, and he is a member of the Mexican Research System. He is the author of more than 60 papers in journals and conferences. His main research interests include Data Science, Social Network Analysis, Agent Theory, and Natural Language Processing, applied to Scientometrics and Learning Analytics.

## ACKNOWLEDGEMENTS

The authors would like to thank Tecnologico de Monterrey and the Living Lab & Data Hub of the Institute for the Future of Education for the data provided for this research (Data Request 202308DHR15).
S Ramos-Pulido thanks Tecnologico de Monterrey and Conahcyt for providing the Ph.D. scholarship.

## REFERENCES

[1] M. Atif, F. Anwer, F. Talib, R. Alam, and F. Masood, "Analysis of machine learning classifiersfor predicting diabetes mellitus in the preliminary stage,"Int J ArtifIntell ISSN, vol. 12, no. 3, pp. 1302-1311, 2023.
[2] N. Dietz, V. Jaganathan, V. Alkin, J. Mettille, M. Boakye, and D. Drazin, "Machine learning in clinical diagnosis,prognostication, and management of acute traumatic spinal cord injury (sci): A systematic review," Journalof Clinical Orthopaedics and Trauma, vol. 35, pp. 102046, 2022.
[3] P. Dixit and S. Silakari, "Deep learning algorithms for cybersecurity applications: A technological and statusreview,"Computer Science Review, vol. 39, pp. 100317, 2021.
[4] M. D. Genemo, "Suspicious activity recognition for monitoring cheating in exams,"Proceedings of the Indian National Science Academy, vol. 88, no. 1, pp. 1–10, 2022.
[5] A. Kamilaris and F. X. Prenafeta-Boldu´, "Deep learning in agriculture: A survey, Computers and electronics in agriculture," vol. 147, pp. 70–90, 2018.

[6] S Wang, S. Jin, D. Deng, and C. Fernandez, “A critical review of online battery remaining usefullifetime prediction methods,” Frontiers in Mechanical Engineering, vol. 7, pp. 719718, 2021.
[7] B. Efron and T. Hastie, “Computer age statistical inference, student edition: algorithms, evidence, and data science.” Cambridge University Press, vol. 6, 2021.
[8] L. Goel and J. Nagpal, “A systematic review of recent machine learning techniques for plant disease identification and classification,” IETE Technical Review, vol. 40, no. 3, pp. 423-439, 2022.
[9] J. Hegde and B. Rokseth, “Applications of machine learning methods for engineering risk assessment–a review,” Safety science, vol. 122, pp. 104492, 2020.
[10] H. Sarker, “Machine learning: Algorithms, real-world applications, and research directions,”SN computer science, vol. 2, no. 3, pp. 160, 2021.
[11] M. M. Taye, “Understanding of machine learning with deep learning: Architectures, workflow, applicationsand future directions,” Computers, vol. 12, no. 5, pp. 91, 2023.
[12] Sharifani and M. Amini, “Machine learning and deep learning: A review of methods and applications,” World Information Technology and Engineering Journal, vol. 10, no. 07, pp. 3897–3904, 2023.
[13] Y. Ma and G. Guo, “Support Vector Machines Applications.”Springer, vol. 649, 2014.
[14] Y.-C. Chang, K.-H. Chang, and G.-J. Wu, “Application of extreme gradient boosting trees in the construction of credit risk assessment models for financial institutions,” Applied Soft Computing, vol. 73, pp. 914–920, 2018.
[15] S. Ramos-Pulido, N. Hernández-Gress, and G. Torres-Delgado, “Analysis of soft skills and job level with data science: A case for graduates of a private university,”Informatics, vol. 10, no. 1, pp. 23, 2023.
[16] D. Vassallo, V. Vella, and J. Ellul, “Application of gradient boosting algorithms for anti-money laundering incryptocurrencies,”SN Computer Science, vol. 2, pp. 1–15, 2021.
[17] A. C. Muller and S. Guido, “Introduction to Machine Learning with Python: A Guide for Data Scientists.” O’Reilly Media, Inc., 2016.
[18] C M. Bishop and N. M. Nasrabadi, “Pattern Recognition and Machine Learning.” Springer, 2006.
[19] Y. Zhang and A. Haghani, “A gradient boosting method to improve travel time prediction,” Transportation Research Part C: Emerging Technologies, vol. 58, pp. 308–324, 2015.
[20] I. Kuncheva, “Combining pattern classifiers: methods and algorithms.” John Wiley & Sons, 2014.
[21] Gori and A. Tesi, “On the problem of local minima in backpropagation,” IEEE Transactions on Pattern Analysis and Machine Intelligence, vol. 14, no. 1, pp. 76–86, 1992.
[22] A.Natekin and A. Knoll, “Gradient boosting machines, a tutorial”, Frontiers in neurorobotics,vol. 7, 2013.
[23] H. Poulard and D. Estéve, “A convergence theorem for barycentric correction procedure,”Soumisa Neural Computation, 1995.
[24] H. Poulard and S. Labreche, “A new unit learning algorithm,” ipi, vol. 10, pp. i2I1, 1995.
[25] E. Bisong, Google Colaboratory.Berkeley, CA: Apress, pp. 59–64, 2019.
[26] T. de Monterrey, “Impacto economico y social global de las y los egresados del exatec en 80 años de historia,” 2023. [Online]. Available: https://egresados.exatec.tec.mx/estudioqs80.
[27] Ring and B. M. Eskofier, “An approximation of the gaussian rbf kernel for efficient classificationwith svms,” Pattern Recognition Letters, vol. 84, pp. 107–113, 2016